\documentclass[runningheads]{llncs}
\usepackage{graphicx}
\usepackage{amsmath,amssymb} % define this before the line numbering.
\usepackage{color}
\usepackage{times}
\usepackage{epsfig}
\usepackage{dsfont}
\usepackage{siunitx}
\usepackage{subfigure}
\usepackage{multirow}
\usepackage{caption}
\usepackage{booktabs}       % professional-quality tables
\usepackage{standalone}
\usepackage{xspace}
\usepackage{comment}
\usepackage{hyperref}
\usepackage{harpoon}

\newcommand{\etal}{\textit{et al}.}
\newcommand{\ie}{\textit{i}.\textit{e}.,~}
\newcommand{\eg}{\textit{e}.\textit{g}.,~}
\newcommand{\myparagraph}[1]{\vspace{6pt}\noindent{\bf #1}}
\newcommand{\beginsupplement}{%
        \setcounter{table}{0}
        \renewcommand{\thetable}{S\arabic{table}}%
        \setcounter{figure}{0}
        \renewcommand{\thefigure}{S\arabic{figure}}%
     }
     
\newcommand{\nUniqueVideos}{6,984\xspace}

\newcommand{\nHours}{77\xspace}

\newcommand{\nVideosTraining}{5,588\xspace}
\newcommand{\nVideosValidation}{698\xspace}
\newcommand{\nVideosTest}{698\xspace}

\begin{document}
\title{Textual Explanations for Self-Driving Vehicles} 
% Replace with your title

\titlerunning{Textual Explanations for Self-Driving Vehicles}
% Replace with a meaningful short version of your title
%

\author{Jinkyu Kim\inst{1} \and
Anna Rohrbach\inst{1,2} \and 
Trevor Darrell\inst{1} \and 
John Canny\inst{1} \and
Zeynep Akata\inst{2,3}
}
%
%Please write out author names in full in the paper, i.e. full given and family names. 
%If any authors have names that can be parsed into FirstName LastName in multiple ways, please include the correct parsing, in a comment to the volume editors:
%\index{Lastnames, Firstnames}
%(Do not uncomment it, because you may introduce extra index items if you do that, we will use scripts for introducing index entries...)
\authorrunning{J. Kim, A. Rohrbach, T. Darrell, J. Canny, Z. Akata}
% Replace with shorter version of the author list. If there are more authors than fits a line, please use A. Author et al.
%

\institute{EECS, University of California, Berkeley CA 94720, USA
\email{\{jinkyu.kim,anna.rohrbach,trevordarrell,canny\}@berkeley.edu}\\ \and 
MPI for Informatics, Saarland Informatics Campus, 66123 Saarbr{\"u}cken, Germany \and 
AMLab, University of Amsterdam, 1098 XH Amsterdam, Netherlands\\
\email{z.akata@uva.nl}}
\maketitle              % typeset the header of the contribution
\begin{abstract}
Deep neural perception and control networks have become key components of self-driving vehicles. User acceptance is likely to benefit from easy-to-interpret textual explanations which allow end-users to understand what triggered a particular behavior. Explanations may be triggered by the neural controller, namely \emph{introspective explanations}, or informed by the neural controller's output, namely \emph{rationalizations}. We propose a new approach to introspective explanations which consists of two parts. First, we use a visual (spatial) attention model to train a convolutional network end-to-end from images to the vehicle control commands, \ie acceleration and change of course. The controller's attention identifies image regions that potentially influence the network's output. Second, we use an attention-based video-to-text model to produce textual explanations of model actions. The attention maps of controller and explanation model are aligned so that explanations are grounded in the parts of the scene that mattered to the controller. We explore two approaches to attention alignment, strong- and weak-alignment. Finally, we explore a version of our model that generates rationalizations, and compare with introspective explanations on the same video segments. 
We evaluate these models on a novel driving dataset with ground-truth human explanations, the Berkeley DeepDrive eXplanation (BDD-X) dataset. Code is available at \url{https://github.com/JinkyuKimUCB/explainable-deep-driving}

\end{abstract}
\section{Introduction}
Deep neural networks are an effective tool~\cite{bojarski2016end,xu2016end} to learn vehicle controllers for self-driving cars in an end-to-end manner. Despite their effectiveness as function estimators, DNNs are typically cryptic black-boxes. There are no explainable states or labels in such a network, and representations are fully distributed as sets of activations. Explainable models that make deep models more transparent are important for a number of reasons: (i) user acceptance -- self-driving vehicles are a radical technology for users to accept, and require a very high level of trust, (ii) understanding and extrapolation of vehicle behavior -- users ideally should be able to anticipate what the vehicle will do in most situations, (iii) effective communication -- they help user communicate preferences to the vehicle and vice versa. 

Explanations can be either {\em{rationalizations}} -- explanations that justify the system's behavior in a post-hoc manner, or {\em{introspective explanations}} -- explanations that are based on the system's internal state. Introspective explanations represent {\em causal} relationships between the system's input and its behavior, and address all the goals above. Rationalizations can address acceptance, (i) above, but are less helpful with (ii) understanding the causal behavior of the model or (iii) communication which is grounded in the vehicle's internal state (known as theory of mind in human communication). 

One way of generating introspective explanations is via visual attention~\cite{xu2015show,kim2017interpretable}. Visual attention filters out non-salient image regions, and image areas inside the attended region have potential causal effect on the output (those outside cannot). As shown in \cite{kim2017interpretable}, additional salience filtering can be applied so that the attention map shows only regions that causally affect the output. Visual attention constrains the reasons for the controllers actions but does not \eg tie specific actions to specific input regions \eg ``the vehicle slowed down because the light controlling the intersection is red''. It is also likely to be less convenient for passengers to replay the attention map vs. a (typically on-demand) speech presentation of a textual explanation. 

%------------------ PLACED FOR BETTER POSITION
\begin{figure}[!t]
    \begin{center}
        \includegraphics[width=0.9\linewidth]{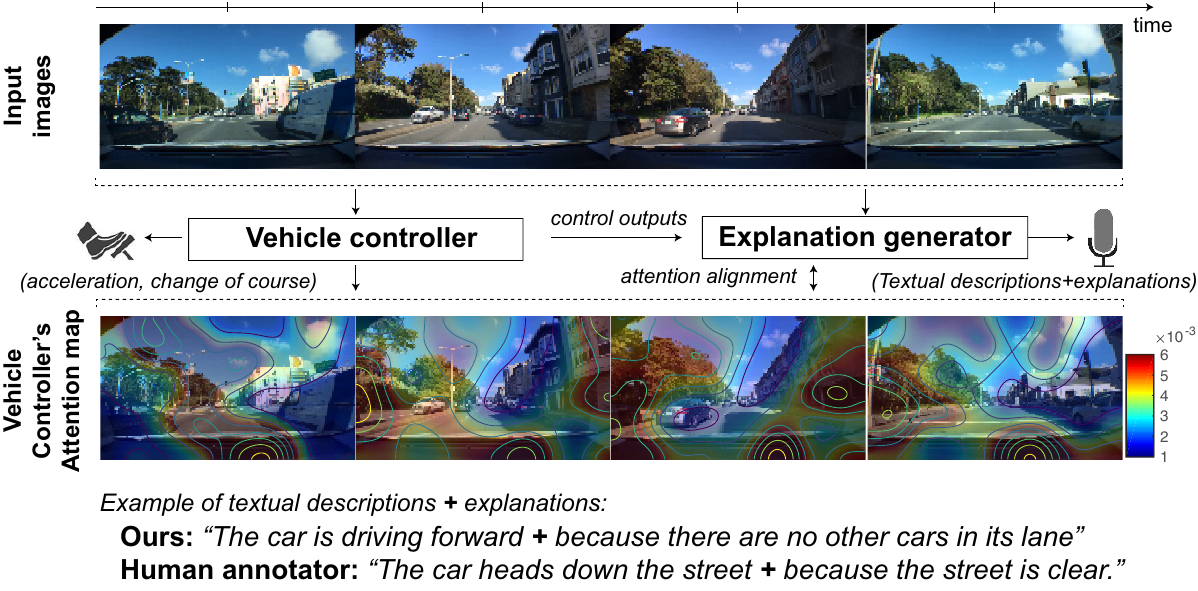}
    \end{center}
    \caption{Our model predicts vehicle’s control commands, \ie an acceleration and a change of course, at each timestep, while an explanation model generates a natural language explanation of the rationales, \eg ``The car is driving forward because there are no other cars in its lane'', and a visual explanation in the form of attention -- attended regions directly influence the textual explanation generation process.}
    \label{fig:quick_overview}
\end{figure}

In this work, we focus on generating textual descriptions and explanations, such as the pair: ``vehicle slows down'' and ``because it is approaching an intersection and the light is red'' as in \autoref{fig:quick_overview}. Natural language has an advantage of being inherently understandable and does not require familiarity with the design of an intelligent system in order to provide useful information. 
In order to train such a model, we collect explanations from human annotators. Our explanation dataset is built on top of another large-scale driving dataset~\cite{xu2016end} collected from dashboard cameras in human driven vehicles. Annotators view the video dataset, compose descriptions of the vehicle's activity and explanations for the actions that the vehicle driver performed. 

Obtaining training data for vehicle explanations is by itself a significant challenge. The ground truth explanations are in fact often rationalizations (generated by an observer rather than the driver), and there are additional challenges with acquiring driver data. But even more than that, it is currently impossible to obtain human explanations of {\em what the vehicle controller was thinking}, \ie a real ground truth.
Nevertheless our experiments show that using {\em attention alignment} between controller and explanation models generally improves the quality of explanations, \ie generates explanations which better match the human rationalizations of the driving videos. 

Our contributions are as follows. (1) We propose an introspective textual explanation model for self-driving cars to provide easy-to-interpret explanations for the behavior of a deep vehicle control network.
(2) We integrate our explanation generator with the vehicle controller by aligning their attentions to ground the explanation, and compare two approaches: attention-aligned explanations and non-aligned rationalizations.
(3) We generated a large-scale Berkeley DeepDrive eXplanation (BDD-X) dataset with over \nUniqueVideos video clips annotated with driving descriptions, \eg ``The car slows down'' and explanations, \eg ``because it is about to merge with the busy highway''. Our dataset provides a new test-bed for measuring progress towards developing explainable models for self-driving cars.

\section{Related Work}
In this section, we review existing work on end-to-end learning for self-driving cars as well as work on visual explanation and justification.

\myparagraph{End-to-End Learning for Self-Driving Cars:}
Most of vehicle controllers for self-driving cars can be divided in two types of approaches~\cite{chen2015deepdriving}: (1) a mediated perception-based approach and (2) an end-to-end learning approach. The mediated perception-based approach depends on recognizing human-designated features, such as lane markings, traffic lights, pedestrians or cars, which generally require demanding parameter tuning for a balanced performance~\cite{paden2016survey}. Notable examples include~\cite{urmson2008autonomous}, \cite{buehler2009darpa}, and \cite{levinson2011towards}. 

As for the end-to-end approaches, recent works~\cite{bojarski2016end,xu2016end} suggest that neural networks can be successfully applied to self-driving cars in an end-to-end manner. Most of these approaches use behavioral cloning that learns a driving policy as a supervised learning problem over observation-action pairs from human driving demonstrations. Among these, \cite{bojarski2016end} present a deep neural vehicle controller network that directly maps a stream of dashcam images to steering controls, while \cite{xu2016end} use a deep neural network that takes input raw pixels and prior vehicle states and predict vehicle's future motion. Despite their potential, the effectiveness of these approaches is limited by their inability to explain the rationale for the system's decisions, which makes their behavior opaque and uninterpretable. In this work, we propose an end-to-end trainable system for self driving cars that is able to justify its predictions visually via attention maps and textually via natural language.

\myparagraph{Visual and Textual Explanations:}
The importance of explanations for an end-user has been studied from the psychological perspective~\cite{lombrozo2012,lombrozo2006}, showing that humans use explanations as a guide for learning and understanding by building inferences and seeking propositions or judgments that enrich their prior knowledge. They usually seek for explanations to fill the requested gap depending on prior knowledge and goal in question.

In support of this trend, recently explainability has been growing as a field in computer vision and machine learning. Especially, there is a growing interest in introspective deep neural networks. \cite{zeiler2014visualizing} use deconvolution to visualize inner-layer activations of convolutional networks. \cite{lecun2015deep} propose automatically-generated captions for textual explanations of images. \cite{bojarski2016visualbackprop} develop a richer notion of contribution of a pixel to the output. However, a difficulty with deconvolution-style approaches is the lack of formal measures of how the network output is affected by spatially-extended features (rather than pixels). Exceptions to this rule are attention-based approaches. \cite{kim2017interpretable} propose attention-based approach with causal filtering that removes spurious attention blobs. However, it is also important to be able to justify the decisions that were made and explain why they are reasonable in a human understandable manner, \ie a natural language. For an image classification problem, \cite{hendricks2016generating,hendricks2018grounding} used an LSTM \cite{hochreiter1997lstm} caption generation model that generates textual justifications for a CNN model. \cite{park2016attentive} combine attention-based model and a textual justification system to produce an interpretable model. To our knowledge, ours is the first attempt to justify the decisions of a real-time deep controller through a combination of attention and natural language explanations on a stream of images.

\section{Explainable Driving Model}
\begin{figure*}[!t]
    \begin{center}
        \includegraphics[width=\linewidth]{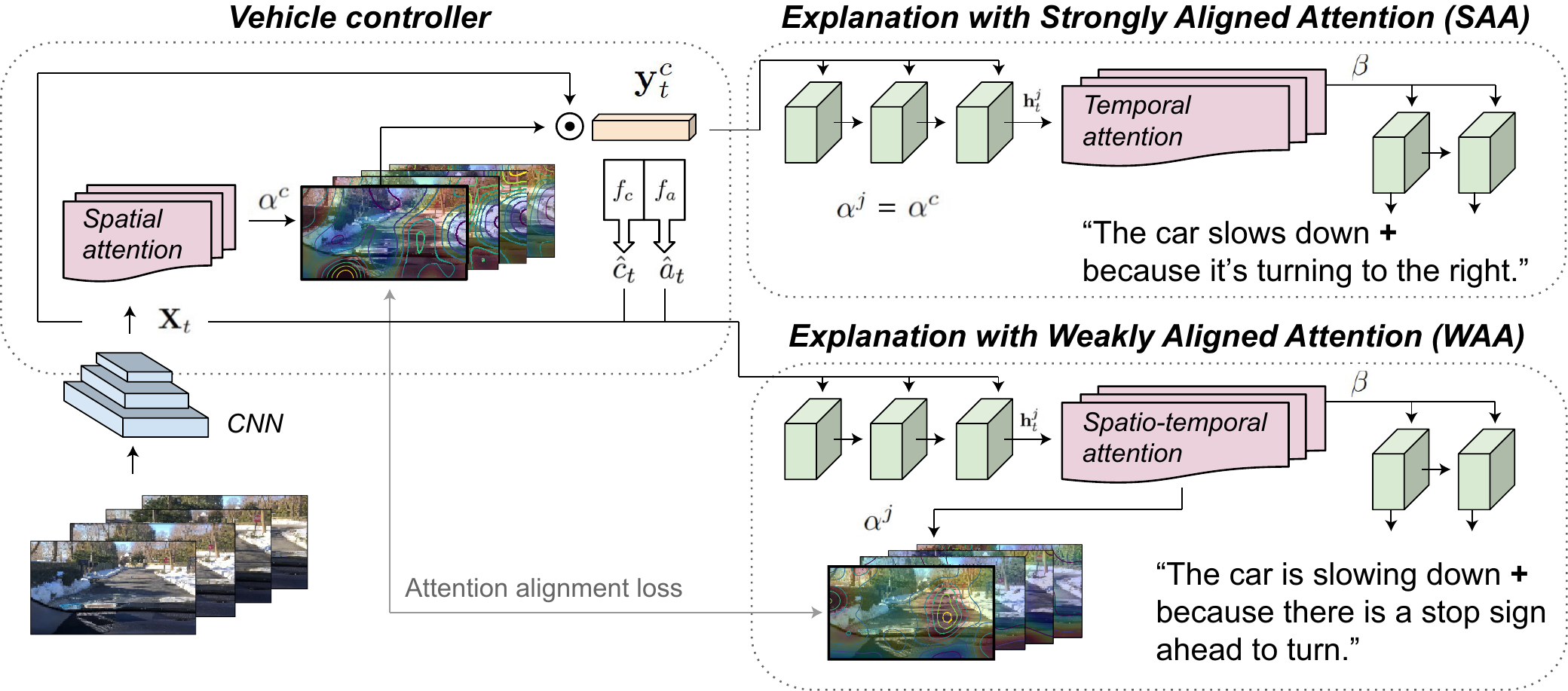}
    \end{center}
    \caption{Vehicle controller generates spatial attention maps $\alpha^c$ for each frame, predicts acceleration and change of course ($\hat{c}_t, \hat{a}_t$) that condition the explanation. Explanation generator predicts temporal attention across frames ($\beta$) and a spatial attention in each frame ($\alpha^j$). SAA uses $\alpha^c$, WAA enforces a loss between $\alpha^j$ and $\alpha^c$.}
    \label{fig:overview}
\end{figure*}

In this paper, we propose a driving model that explains how a driving decision was made both (i) by visualizing image regions where the decision maker attends to and (ii) by generating a textual description and explanation of what has triggered a particular driving decision, \eg ``the car continues (description) because traffic flows freely (explanation)''. As we summarize in~\autoref{fig:overview}, our model involves two parts: (1) a {\em{Vehicle controller}}, which is trained to learn human-demonstrated vehicle control commands, \eg an acceleration and a change of course; our controller uses a visual (spatial) attention mechanism that identifies potentially influential image regions for the network's output; (2) a {\em{Textual explanation generator}}, which generates textual descriptions and explanations controller behavior. The key to the approach is to align the attention maps.

\myparagraph{Preprocessing.}
Our model is trained to predict two vehicle control commands, \ie an acceleration and a change of course. At each time $t$, an acceleration, $a_t$, is measured by taking the derivative of speed measurements, and a change of course, $c_t$, is computed by taking a difference between a current vehicle's course and a smoothed value by using simple exponential smoothing method~\cite{hyndman2008forecasting}. We provide details in supplemental material. To reduce computational burden, we down-sample to 10Hz and reduce the input dimensionality by resizing raw images to a 90$\times$160$\times$3 image with nearest-neighbor scaling algorithm. Each image is then normalized by subtracting the mean from the raw input pixels and dividing by its standard deviation. This preprocessing is applied to the latest 4 frames, which are then stacked to produce the final input to the neural network.

\myparagraph{Convolutional Feature Encoder.}
We use a convolutional neural network to encode the visual information into a set of visual feature vectors at time $t$, \ie convolutional feature cube 
${\bf{X}}_{t} = \{ {\bf{x}}_{t,1}, {\bf{x}}_{t,2},\ldots, {\bf{x}}_{t,l} \} $
where ${\bf{x}}_{t,i}\in{\cal{R}}^{d}$ for $i\in\{1,2,\ldots,l\}$ and $l$ is the number of different spatial regions of the given input. Each feature vector contains a high-level description of objects present in a certain input region. This allows us to focus selectively on different regions of the given image by choosing a subset of these feature vectors.
We use a five-layered convolutional network as in~\cite{bojarski2016end,kim2017interpretable} and omit max-pooling layers to prevent spatial information loss~\cite{lee2009convolutional}. The output is a three-dimensional feature cube ${{\bf{X}}_t}$ and the feature block has the size $w$$\times$$h$$\times$$d$ at each time $t$.

\subsection{Vehicle Controller}
Our vehicle controller is trained in an end-to-end manner. Given a stream of dashcam images and the vehicle's (current) sensor measurements, \eg speed, the controller predicts the acceleration and the change of course at each timestep.
We utilize a deterministic soft attention mechanism that is trainable by standard back-propagation methods. The soft attention mechanism applies attention weights multiplicatively to the features and additively pools the results through the maps $\pi$. 
Our model feeds the context vectors ${\bf{y}}_t^{c}$ produced by the controller map $\pi^{c}$ to the controller LSTM:
\begin{equation}
{\bf{y}}_t^{c} = \pi^{c}(\{\alpha_{t,i}^{c}\}, \{{\bf{x}}_{t,i}\}) = \sum_{i=1}^l\alpha_{t,i}^{c}{\bf{x}}_{t,i}
\end{equation}
where $i=\{1,2,\dots,l\}$. $\alpha_{t,i}^{c}$ is an attention weight map output by a spatial softmax and satisfies $\sum_{i}{\alpha_{t,i}^{c}}=1$. These attention weights can be interpreted as the probability over $l$ convolutional feature vectors. A location with a high attention weight is salient for the task (driving). The attention model $f_{\textnormal{attn}}^{c}({\bf{X}}_{t},{\bf{h}}_{t-1}^{c})$ is conditioned on the previous LSTM state ${\bf{h}}_{t-1}^{c}$, and the current feature vectors ${\bf{X}}_{t}$. It comprises a fully-connected layer and a spatial softmax to yield normalized $\{\alpha_{t,i}^{c}\}$.

The outputs of the vehicle controller are the vehicle's acceleration $\hat{a}_{t}$ and the change of course $\hat{c}_{t}$. To this end, we use additional multi-layer fully-connected blocks with ReLU non-linearities, denoted by $f_{\textnormal{a}}({\bf{y}}_t^{c}, {\bf{h}}_t^{c})$ and $f_{\textnormal{c}}({\bf{y}}_t^{c}, {\bf{h}}_t^{c})$. We also add the entropy $H$ of the attention weight to the objective function:
\begin{equation}
\mathcal{L}_\textnormal{c} = \sum_{t}\big((a_{t}-\hat{a}_{t})^{2}+(c_{t}-\hat{c}_{t})^{2} + \lambda_\textnormal{c}H(\alpha_{t}^{c})\big)
\end{equation}
The entropy is computed on the attention map as though it were a probability distribution. Minimizing loss corresponds to minimizing entropy. Low entropy attention maps are sparse and emphasize relatively few regions. We use a hyperparameter $\lambda_\textnormal{c}$ to control the strength of the entropy regularization term.

\subsection{Attention Alignments}~\label{ss:alignment}
The controller attention map provides input regions that the network attends to, and these regions have a direct influence on the network's output. Thus, to yield ``introspective'' explanation, we argue that the agent must attend to those areas. For example, if a vehicle controller predicts ``acceleration'' by detecting a green traffic light, the textual justification must mention this evidence, \eg ``because the light has turned green''. 
Here, we explain two approaches to align the vehicle controller and the textual justifier such that they look at the same input regions.

\myparagraph{Strongly Aligned Attention (SAA):} A consecutive set of spatially attended input regions, each of which is encoded as a context vector ${\bf{y}}_t^{c}$ by the vehicle controller, can be directly used to generate a textual explanation (see Figure~\ref{fig:overview}, right-top). Thus, models share a single layer of an attention. 
As we detail in Section~\ref{ss:justification}, 
our explanation module uses {\em temporal} attention with weights $\beta$ to the controller context vectors $\{{\bf{y}}_t^j,t=1,\ldots\}$ directly, and thus allows flexibility in output tokens relative to input samples.

\myparagraph{Weakly Aligned Attention (WAA):} Instead of directly using vehicle controller's attention, an explanation generator can have its own spatial attention network (see Figure~\ref{fig:overview}, right-bottom). A loss, \ie the Kullback-Leibler divergence ($D_{\textnormal{KL}}$), between the two attention maps makes the explanation generator refer to the salient objects:
\begin{equation}
 \mathcal{L}_{a} = \lambda_{a} \sum_{t}D_{\textnormal{KL}}(\alpha_t^{c}||\alpha_t^{j})=\lambda_{a}\sum_{t}\sum_{i=1}^{l}\alpha_{t,i}^{c}(\log \alpha_{t,i}^{c} - \log \alpha_{t,i}^{j})
\end{equation}
where $\alpha^{c}$ and $\alpha^{j}$ are the attention maps generated by the vehicle controller and the explanation generator model, respectively. We use a hyperparameter $\lambda_{a}$ to control the strength of the regularization term.

\subsection{Textual Explanation Generator}~\label{ss:justification}
Our textual explanation generator takes sequence of video frames of variable length and generates a variable-length description/explanation. Descriptions and explanations are typically part of the same sentence in the training data but are annotated with a separator. In training and testing we use a synthetic separator token \verb+<sep>+ between description and explanation, but treat them as a single sequence. The explanation LSTM predicts the description/explanation sequence and outputs per-word softmax probabilities. 

The source of context vectors for the description generator depends on the type of alignment between attention maps. For weakly aligned attention or rationalizations, the explanation generator creates its own spatial attention
map $\alpha^j$ at each time step $t$. This map includes a loss against the controller attention map for weakly-aligned attention, but has no such loss when generating rationalizations. The attention map $\alpha^j$ is applied to the CNN output yielding context vectors ${\bf{y}}_t^j$. 

Our textual explanation generator explains the rationale behind the driving model, and thus we argue that a justifier needs the outputs from the vehicle motion predictor as an input. We concatenate a tuple $(\hat{a}_{t}, \hat{c}_{t})$ with a spatially-attended context vector ${\bf{y}}_t^{j}$ and ${\bf{y}}_t^{c}$ respectively for weakly and strongly aligned attention approaches. This concatenated vector is then used to update the LSTM for a textual explanation generation.

The explanation module applies {\em temporal} attention with weights $\beta$ to either the controller context vectors directly $\{{\bf{y}}_t^c,t=1,\ldots\}$ (strong alignment), or to the explanation vectors $\{{\bf{y}}_t^j,t=1,\ldots\}$ (weak alignment or rationalization). Such input sequence attention is common
in sequence-to-sequence models and allows flexibility in output tokens relative
to input samples \cite{bahdanau2014neural}. The result of temporal attention application is (dropping the $c$ or $j$ superscripts on ${\bf{y}}$):
\begin{equation}
{\bf{z}}_k = \pi(\{\beta_{k,t}\}, \{{\bf{y}}_{t}\}) = \sum_{t=1}^{T}\beta_{k,t}{\bf{y}}_{t}
\end{equation}
where $\sum_{t}{\beta_{k,t}}=1$. The weight $\beta_{k,t}$ at each time $k$ (for sentence generation) is computed by an attention model $f_{\textnormal{attn}}^{e}(\{{\bf{y}}_{t}\},{\bf{h}}_{k-1}^{e})$, which is similar to the spatial attention as we explained in previous section (see supplemental material for details). 

To summarize, we minimize the following negative log-likelihood (for training our justifier) as well as vehicle control estimation loss $\mathcal{L}_{c}$ and attention alignment loss $\mathcal{L}_{a}$:
\begin{equation}
\mathcal{L} = \mathcal{L}_{c}+\mathcal{L}_{a}-\sum_{k}\log p({\bf{o}}_k|{\bf{o}}_{k-1}, h_{k}^{e}, {\bf{z}}_{k})
\end{equation}

\section{Berkeley DeepDrive eXplanation Dataset (BDD-X)}
\begin{figure}[!t]
    \begin{center}
        \includegraphics[width=\linewidth]{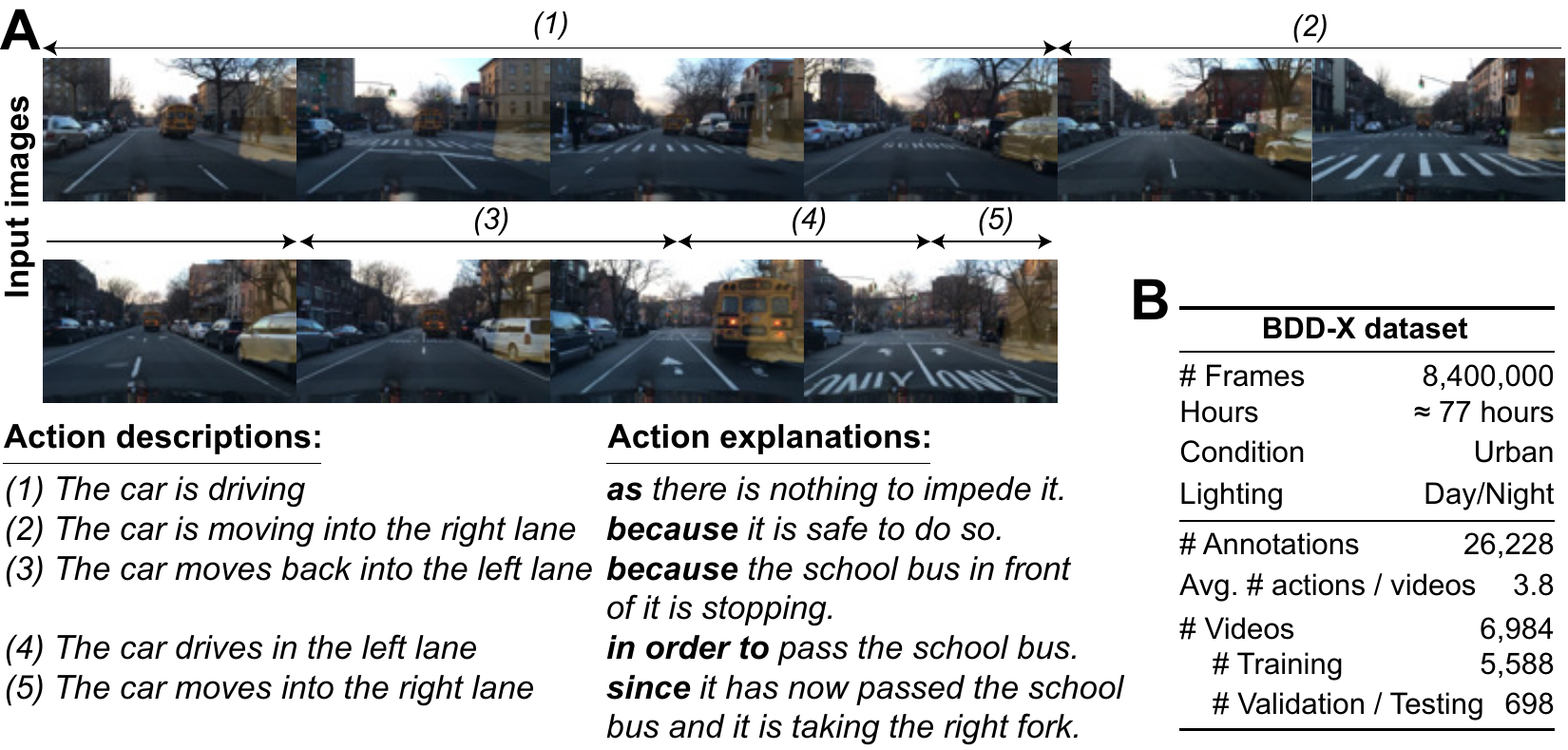}
    \end{center}
    \caption{(A) Examples of input frames and corresponding human-annotated action description and justification of how a driving decision was made. For visualization, we sample frames at every two seconds. (B) BDD-X dataset details. Over \nHours hours of driving with time-stamped human annotations for action descriptions and justifications.} 
    \label{fig:dataset}
\end{figure}

In order to effectively generate and evaluate textual driving rationales we have collected textual justifications for a subset of the Berkeley Deep Drive (BDD) dataset~\cite{xu2016end}. This dataset contains videos, approximately 40 seconds in length, captured by a dashcam mounted behind the front mirror of the vehicle. Videos are mostly captured during urban driving in various weather conditions, featuring day and nighttime. The dataset also includes driving on other road types, such as residential roads (with and without lane markings), and contains all the typical driver's activities such as staying in a lane, turning, switching lanes, etc. Alongside the video data, the dataset provides a set of time-stamped sensor measurements, such as vehicle's velocity, course, and GPS location. For sensor logs unsynchronized with the time-stamps of video data, we use the estimates of the interpolated measurements. 

In order to increase trust and reliability, the machine learning system underlying self driving cars should be able to explain why at a certain time they make certain decisions. Moreover, a car that justifies its decision through natural language would also be user friendly. Hence, we populate a subset of the BDD dataset with action description and justification for all the driving events along with their timestamps. We provide examples from our \emph{Berkeley Deep Drive eXplanation (BDD-X)} dataset in Figure~\ref{fig:dataset} (A). 

\myparagraph{Annotation.} We provide a driving video and ask a human annotator in Amazon Mechanical Turk to imagine herself being a driving instructor. Note that we specifically select human annotators who are familiar with US driving rules. The annotator has to describe \emph{what} the driver is doing (especially when the behavior changes) and \emph{why}, from a point of view of a driving instructor. Each described action has to be accompanied with a start and end time-stamp. The annotator may stop the video, forward and backward through it while searching for the activities that are interesting and justifiable. 

To ensure that the annotators provide us the driving rationales as well as descriptions, we require that they separately enter the \emph{action description} and the \emph{action justification}: \eg \emph{``The car is moving into the left lane''} and \emph{``because the school bus in front of it is stopping.''}. In our preliminary annotation studies, we found that giving separate annotation boxes is helpful for the annotator to understand the task and perform better.

\myparagraph{Dataset Statistics.} Our dataset (see Figure~\ref{fig:dataset} (B)) is composed of over \nHours hours of driving within \nUniqueVideos videos. The videos are taken in diverse driving conditions, \eg day/night, highway/city/countryside, summer/winter etc. On an average of 40 seconds, each video contains around 3-4 actions, \eg speeding up, slowing down, turning right etc., all of which are annotated with a description and an explanation. Our dataset contains over $26K$ activities in over $8.4M$ frames. We introduce a training, a validation and a test set, containing \nVideosTraining, \nVideosValidation and \nVideosTest videos, respectively.

\myparagraph{Inter-human agreement.} 
Although we cannot have access to the internal thought process of the drivers, one can infer the reason behind their actions using the visual evidence of the scene. Besides, it would be challenging to setup the data collection process which enables drivers to report justifications for all their actions, if at all possible. We ensure the high quality of the collected annotations by relying on a pool of qualified workers (\ie they pass a qualification test) and selective manual inspection. 

Further, we measure the inter-human agreement on a subset of 998 training videos, each of which has been annotated by two different workers. Our analysis is as follows. In 72\% of videos the number of annotated intervals differs by less than 3. The average temporal $IoU$ across annotators is $0.63$ ($SD=0.21$). When $IoU>0.5$ the CIDEr score across action descriptions is 142.60, across action justifications it is 97.49 (random choice: 39.40/28.39, respectively). When $IoU>0.5$ and action descriptions from two annotators are identical (165 clips\footnote[1]{The number of video intervals (not full videos), where the provided action descriptions (not explanations) are identical (common actions \eg ``the car slows down'').}) the CIDEr score across justifications is 200.72, while a strong baseline, selecting a justification from a different video with the same action description, results in CIDEr score 136.72. These results show an agreement among annotators and relevance of collected action descriptions and justifications.

\myparagraph{Coverage of justifications.}
BDD-X dataset has over 26k annotations (77 hours) collected from a substantial random subset of large-scale crowd-sourced driving video dataset, which consists of all the typical driver’s activities during urban driving. The vocabulary of training action descriptions and justifications is 906 and 1,668 words respectively, suggesting that justifications are more diverse than descriptions. Some of the common actions are (frequency decreasing): moving forward, stopping, accelerating, slowing, turning, merging, veering, pulling [in]. Justifications cover most of the relevant concepts: traffic signs/lights, cars, lanes, crosswalks, passing, parking, pedestrians, waiting, blocking, safety etc.

\section{Results and Discussion}
% %----------- PLACE FOR BETTER POSITION
\begin{figure}[!t]
\begin{center}
\includegraphics[width=\linewidth]{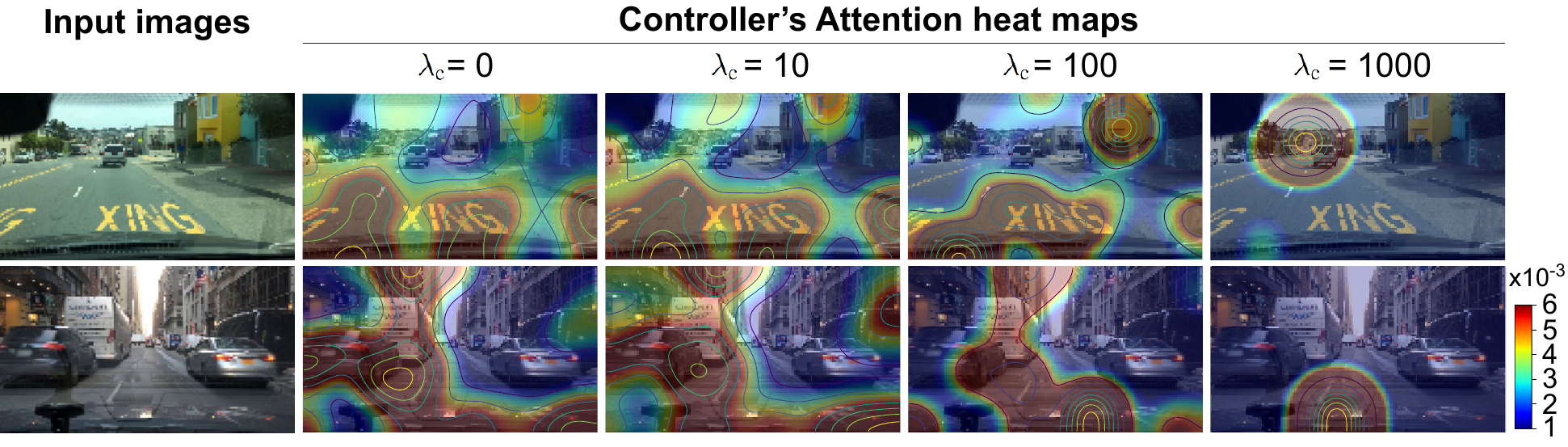}
\end{center}
   \caption{Vehicle controller’s attention maps in terms of four different entropy regularization coefficient  $\lambda_{c}$=\{0,10,100,1000\}. Red parts indicate where the model pays more attention. Higher value of $\lambda_{c}$ makes the attention maps sparser. We observe that sparser attention maps improves the performance of generating textual explanations, while control performance is slightly degraded.
} 
\label{fig:entReg}
\end{figure}

Here, we first provide our training and evaluation details, then make a quantitative and qualitative analysis of our vehicle controller and our textual justifier.

\myparagraph{Training and Evaluation Details.}
As the convolutional feature encoder, we use 5-layer CNN~\cite{bojarski2016end} that produces a 12$\times$20$\times$64-dimensional convolutional feature cube from the last layer. The controller following the CNN has 5 fully connected layers (\ie \#hidden dims: 1164, 100, 50, 10, respectively), which predict the acceleration and the change of course, and is trained end-to-end from scratch. Using other more expressive networks may give a performance boost over our base CNN configuration, but these explorations are out of our scope. Given the obtained convolutional feature cube, we first train our vehicle controller, and then the explanation generator (single layer LSTM unless stated otherwise) by freezing the control network. For training, we use Adam optimizer~\cite{kingma2014adam} and dropout~\cite{srivastava2014dropout} of 0.5 at hidden state connections and Xavier initialization~\cite{glorot2010understanding}. The standard dataset is split as 80\% (\nVideosTraining videos) as the training set, 10\% (\nVideosTest videos) as the test, and 10\% (\nVideosValidation videos) as the validation set. Our model takes less than a day to train on a single NVIDIA Titan X GPU.

For evaluating the vehicle controller we use the mean absolute error (lower is better) and the distance correlation (higher is better) and for the justifier we use BLEU~\cite{papineni2002bleu}, METEOR~\cite{lavie2005meteor}, and CIDEr-D~\cite{vedantam2015cider}, as well as human evaluation. The former metrics are widely used for the evaluation of video and image captioning models automatically against ground truth.

{
\setlength{\tabcolsep}{4pt}
\renewcommand{\arraystretch}{1.3} 
\begin{table}[t]
	\begin{center}
    	\resizebox{\linewidth}{!}{%
    	\begin{tabular}{@{}lrcccc@{}} \toprule
        	\multirow{2}{*}{Model} & \multirow{2}{*}{$\lambda_{c}$}& \multicolumn{2}{c}{Mean of absolute error (MAE)} & \multicolumn{2}{c}{Mean of distance correlation}\\ \cmidrule{3-6}
        	& & Acceleration (m/s$^2$) & Course (degree) & Acceleration (m/s$^2$) & Course (degree) \\ \midrule
            CNN+FC~\cite{bojarski2016end}$^{\dagger}$ &- & 6.92 [7.50] & 12.1 [19.7] & 0.17 [0.15] & 0.16 [0.14] \\
            CNN+FC~\cite{bojarski2016end}+P & -& 6.09 [7.73] & 6.74 [14.9] &  0.21 [0.18] & 0.39 [0.33] \\
            CNN+LSTM+Attention~\cite{kim2017interpretable}$^{\dagger}$& -& 6.87 [7.44] & 10.2 [18.4] & 0.19 [0.16] & 0.22 [0.18] \\ \midrule
            CNN+LSTM+Attention+P (Ours)  & 1000&5.02 [6.32] & 6.94 [15.4] & 0.65 [0.25] & 0.43 [0.33] \\ 
            CNN+LSTM+Attention+P (Ours)  & 100& 2.68 [3.73] & 6.17 [14.7] & 0.78 [0.28] & 0.43 [0.34] \\ 
            CNN+LSTM+Attention+P (Ours)  & 10 & 2.33 [3.38] & 6.10 [14.7] & 0.81 [0.27] & 0.46 [0.35] \\
            CNN+LSTM+Attention+P (Ours)  & 0 & {\bf{2.29 [3.33]}} & {\bf{6.06 [14.7]}} & {\bf{0.82 [0.26]}} & {\bf{0.47 [0.35]}} \\ \bottomrule
        \end{tabular}}
    \end{center}
    \caption{Comparing variants of our vehicle controller with different values of the entropy regularization coefficient $\lambda_{c}$=$\{0,10,100,1000\}$ and the state-of-the-art. High value of $\lambda_{c}$ produces low entropy attention maps that are sparse and emphasize relatively few regions. $^{\dagger}$: Models use a single image frame as an input. The standard deviation is in braces. \textit{Abbreviation:} FC (fully connected layer), P (prior inputs)}
   \label{Table:predictor_performance}
\end{table}
}

\subsection{Evaluating Vehicle Controller}

We start by quantitatively comparing variants of our vehicle controller and the state of the art, which include variants of the work by Bojarski~\etal~\cite{bojarski2016end} and Kim~\etal~\cite{kim2017interpretable} in~\autoref{Table:predictor_performance}. Note that these works differ from ours in that their output is the curvature of driving, while our model estimates continuous acceleration and the change of course values. Thus, their models have a single output, while ours estimate both control commands. In this experiment, we replaced their output layer with ours. For a fair comparison, we use an identical CNN for all models.

In this experiment, each model estimates vehicle's acceleration and the change of course.  
Our vehicle controller predicts acceleration and the change of course, which generally requires prior knowledge of vehicle's current state, \ie speed and course, and navigational inputs, especially in urban driving. We observe that the use of the latest four consecutive frames and prior inputs (\ie vehicle's motion measurement and navigational information) improves the control prediction accuracy (see 3rd vs. 7th row), while the use of visual attention also provides improvements (see 1st vs. 3rd row). Specifically, our model without the entropy regularization term (last row) performs the best compared to CNN based approaches~\cite{bojarski2016end} and \cite{kim2017interpretable}. The improvement is especially pronounced for acceleration estimation. 

In \autoref{fig:entReg} we compare input images (first column) and corresponding attention maps for different entropy regularization coefficients $\lambda_{c}$=$\{0,10,100,1000\}$. Red is high attention, blue is low. As we see, higher $\lambda_{c}$ lead to sparser maps. For better visualization, an attention map is overlaid by its contour lines and an input image. 

Quantitatively, the controller performance (error and correlation) slightly degrade as $\lambda_c$ increases and the attention maps become more sparse (see bottom four rows in~\autoref{Table:predictor_performance}). So there is some tension between sparse maps (which are more interpretable), and controller performance. An alternative to regularization, \cite{kim2017interpretable} use causal filtering over the controller's attention maps and achieve about 60\% reduction in ``hot'' attention pixels. 
Causal filtering is desirable for the present work not only to improve sparseness but because after causal filtering, ``hot'' regions necessarily {\em do} have a causal effect on controller behavior, whereas unfiltered attention regions may not. We will explore it in future work.

{
\setlength{\tabcolsep}{4pt}
\renewcommand{\arraystretch}{1.3}
\begin{table}[t]
	\begin{center}
    	\resizebox{\linewidth}{!}{%
    	\begin{tabular}{@{}llcrrcccccc@{}} \toprule
        	\multirow{3}{*}{Type}& \multirow{3}{*}{Model}& \multirow{3}{*}{\parbox{1cm}{\centering Control inputs}}& \multirow{3}{*}{$\lambda_{a}$}& \multirow{3}{*}{$\lambda_{c}$}& \multicolumn{3}{c}{Explanations}& \multicolumn{3}{c}{Descriptions}  \\ 
             & & & & & \multicolumn{3}{c}{(\eg {\em{``because the light is red''}})}& \multicolumn{3}{c}{(\eg {\em{``the car stops''}})} \\ \cmidrule{6-11}
        	 &  &  & & & BLEU-4 & METEOR & CIDEr-D & BLEU-4 & METEOR & CIDEr-D \\ \midrule
            \multirow{3}{*}{} &S2VT~\cite{venugopalan2015sequence} & N& -& -&6.332& 11.19& 53.35& 30.21& 27.53& 179.8\\
            &S2VT~\cite{venugopalan2015sequence}+SA & N& -& -& 5.668& 10.96& 51.37& 28.94& 26.91& 171.3\\ 
            &S2VT~\cite{venugopalan2015sequence}+SA+TA & N& -& -& 5.847& 10.91& 52.74& 27.11& 26.41& 157.0\\ \midrule
            {\em{Rationalization}}&Ours (no constraints)& Y& 0& 0& 6.515 & 12.04 & 61.99& 31.01& 28.64& 205.0\\ \midrule
            \multirow{6}{*}{\parbox{2cm}
            { {\em{Introspective explanation}}}} &Ours (with SAA)& Y& -& 0& 6.998& 12.08& 62.24& {\bf{32.44}}& 29.13& 213.6\\ 
            &Ours (with SAA)& Y& -& 10& 6.760& 12.23& 63.36& 29.99& 28.26& 203.6\\ 
            &Ours (with SAA)& Y& -& 100& 7.074& 12.23& 66.09& 31.84& 29.11& 214.8\\ \cmidrule{2-11}
            &Ours (with WAA)& Y& 10& 0& 6.967& 12.14& 64.19& 32.24& 29.00& {\bf{219.7}}\\
            &Ours (with WAA)& Y& 10& 10 & 6.951& {\bf{12.34}}& 68.56& 30.40& 28.57& 206.6\\
            &Ours (with WAA)& Y& 10& 100 & {\bf{7.281}}& 12.24& {\bf{69.52}}& 32.34& {\bf{29.22}}& 215.8\\ \bottomrule
        \end{tabular}}
    \end{center}
    \caption{Comparing generated and ground truth (columns 6-8) descriptions (\eg ``the car stops'') and explanations (\eg ``because the light is red''). We implement S2VT~\cite{venugopalan2015sequence} and variants with spatial attention (SA) and temporal attention (TA) as a baseline. We tested two different attention alignment approaches, \ie WAA (weakly aligned attention) and SAA (strongly aligned attention), with different combinations of two regularization coefficients: $\lambda_{a}$=$\{0,10\}$ for the attention alignment and $\lambda_{c}$=$\{0,10,100\}$ for the vehicle controller. Rationalization baseline relies on our model (WAA approach) but has no attention alignment. Note that we report all values as a percentage.}
    \label{Table:justifier}
\end{table}
}

\subsection{Evaluating Textual Explanations}
In this section, we evaluate textual explanations against the ground truth explanation using automatic evaluation measures, and also provide human evaluation followed by a qualitative analysis. 

\myparagraph{Automatic Evaluation.}
For state-of-the-art comparison, we implement the S2VT~\cite{venugopalan2015sequence} and its variants. Note that in our implementation S2VT uses our CNN and does not use optical flow features. In~\autoref{Table:justifier}, we report a summary of our experiment validating the quantitative effectiveness of our approach. Rows 5-10 show that best explanation results are generally obtained with weakly-aligned attention. Comparing with row 4, the introspective models all gave higher scores than the rationalization model for explanation generation. Description scores are more mixed, but most of the introspective model scores are higher. As we will see in the next section, our rationalization model focuses on visual saliencies, which is sometimes different from what controller actually ``looks at''. For example, in Figure 5 (5th example), our controller sees the front vehicle and our introspective models generate explanations such as ``because the car in front is moving slowly'', while our rationalization model does not see the front vehicle and generates ``because it's turning to the right''. 

As our training data are human observer annotations of driving videos, and they are not the explanations of drivers, they are post-hoc rationalizations. However, based on the visual evidence, (\eg the existence of a turn right sign explains why the driver has turned right even if we do not have access to the exact thought process of the driver), they reflect typical causes of human driver behavior. The data suggest that grounding the explanations in controller internal state helps produce explanations that better align with human third-party explanations. Biasing the explanations toward controller state (which the WAA and SAA models do) improves their plausibility from a human perspective, which is a good sign. We further analyze human preference in the evaluation below.

{
\setlength{\tabcolsep}{4pt}
\renewcommand{\arraystretch}{1.3}
\begin{table}[t]
	\begin{center}
    	\resizebox{0.9\linewidth}{!}{%
    	\begin{tabular}{@{}llcrrcc@{}} \toprule
        	\multirow{2}{*}{Type}& \multirow{2}{*}{Model}& \multirow{2}{*}{\parbox{1cm}{\centering Control inputs}}& \multirow{2}{*}{$\lambda_{a}$}& \multirow{2}{*}{$\lambda_{c}$}& \multicolumn{2}{c}{Correctness rate}  \\ \cmidrule{6-7}
             & & & & & Explanations& Descriptions \\ \midrule
            {\em{Rationalization}}&Ours (no constraints)& Y& 0& 0& 64.0\% & 92.8\%\\ \midrule
            \multirow{2}{*}{\parbox{2cm}{ {\em{Introspective explanation}}}}&Ours (with SAA)& Y& -& 100& 62.4\%& 90.8\%\\
            &Ours (with WAA)& Y& 10& 100 & {\bf{66.0\%}}& {\bf{93.5\%}}\\ \bottomrule
        \end{tabular}}
    \end{center}
    \caption{Human evaluation of the generated action descriptions and explanations for randomly chosen 250 video intervals. We measure the success rate where at least 2 human judges rate the generated description or explanation with a score 1 (correct and specific/detailed) or 2 (correct).}
    \label{Table:humaneval}
\end{table}
}

\myparagraph{Human Evaluation.}
In our first human evaluation experiment the human judges are only shown the \emph{descriptions}, while in the second experiment they only see the \emph{explanations} (e.g. ``\emph{The car ... because} $<explanation>$''), to exclude the effect of explanations/descriptions on the ratings, respectively. We randomly select 250 video intervals and compare the Rationalization, WAA ($\lambda_{a}$=$10$, $\lambda_{c}$=$100$) and SAA ($\lambda_{c}$=$100$) predictions. The humans are asked to rate a description/explanation on the scale \{1..4\} (1: correct and specific/detailed, 2: correct, 3: minor error, 4: major error). We collect ratings from 3 human judges for each task. Finally, we compute the majority vote, \ie at least 2 out of 3 judges should rate the description/explanation with a score 1 or 2. 

\begin{figure}[t]
\begin{center}
\includegraphics[width=\linewidth]{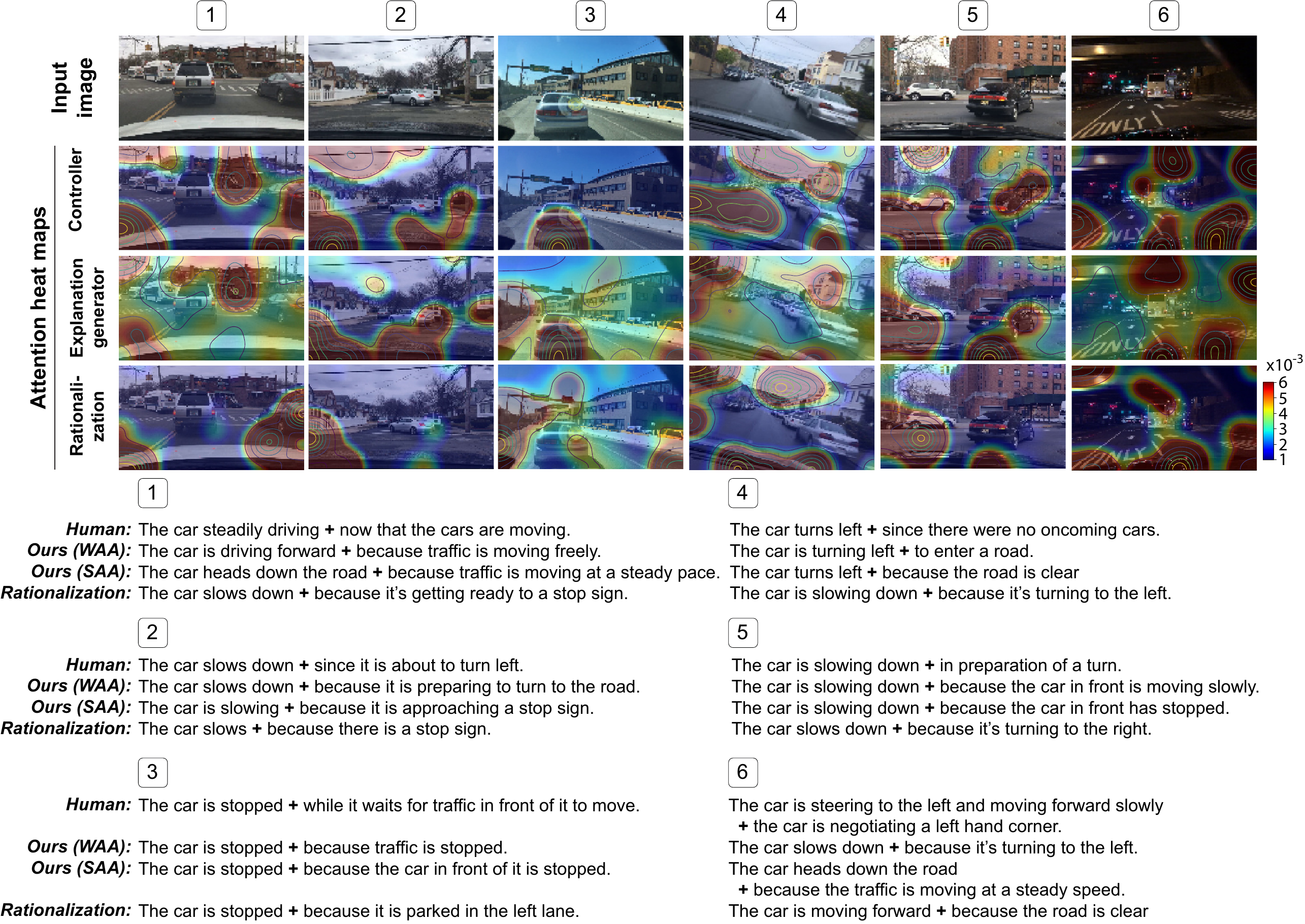}
\end{center}
   \caption{Example descriptions and explanations generated by our model compared to human annotations. We provide (top row) input raw images and attention maps by (from the 2nd row) vehicle controller, textual explanation generator, and rationalization model (Note: ($\lambda_{c}, \lambda_{a}$) = (100,10) and the synthetic separator token is replaced by `+'). 
}
\label{fig:WAA}
\end{figure}

As shown in Table~\ref{Table:humaneval}, our WAA model outperforms the other two, supporting the results above. Interestingly, Rationalization does better than SAA on this subset, according to humans. This is perhaps because the explanation in SAA relies on the exact same visual evidence as the controller, which may include counterfactually important regions (\ie there \emph{could be} a stop sign here), but may confuse the explanation module.

\myparagraph{Qualitative Analysis of Textual Justifier.}
As \autoref{fig:WAA} shows, our proposed textual explanation model generates plausible descriptions and explanations, while our model also provides attention visualization of their evidence. In the first example of \autoref{fig:WAA}, controller sees neighboring vehicles and lane markings, while explanation model generates ``the car is driving forward (description)'' and ``because traffic is moving freely (explanation)''. In~\autoref{fig:WAA}, we also provide other examples that cover common driving situations, such as driving forward (1st example), slowing/stopping (2nd, 3rd, and 5th), and turning (4th and 6th). We also observe that our explanations have significant diversity, \eg they provide various reasons for stopping: red lights, stop signs, and traffic. We provide more diverse examples as supplemental materials.

\section{Conclusion}
We described an end-to-end explainable driving model for self-driving cars by incorporating a grounded introspective explanation model. We showed that (i) incorporation of an attention mechanism and prior inputs improves vehicle control prediction accuracy compared to baselines, (ii) our grounded (introspective) model generates accurate human understandable textual descriptions and explanations for driving behaviors, (iii) attention alignment is shown to be effective at combining the vehicle controller and the justification model, and (iv) our BDD-X dataset allows us to train and automatically evaluate our interpretable justification model by comparing with human annotations.

Recent work~\cite{kim2017interpretable} suggests that causal filtering over attention heat maps can achieve a useful reduction in explanation complexity by removing spurious blobs, which do not significantly affect the output. Causal filtering idea would be worth exploring to obtain causal attention heat maps, which can provide the causal ground of reasoning. Furthermore, it would be beneficial to incorporate stronger perception pipeline, e.g.  object detectors, to introduce more ``grounded'' visual representations and further improve the quality and diversity of the generated explanations. Besides, incorporating driver's eye gaze into our explanation model for mimicking driver's behavior, would be an interesting potential future direction.

\paragraph{Acknowledgements.} This work was supported by DARPA XAI program and Berkeley DeepDrive.

\bibliographystyle{splncs04}
\bibliography{ECCV}

\beginsupplement
\newpage
\title{Supplemental Material:\\Textual Explanations for Self-Driving Vehicles} 
\author{Jinkyu Kim\inst{1} \and
Anna Rohrbach\inst{1,2} \and 
Trevor Darrell\inst{1} \and 
John Canny\inst{1} \and
Zeynep Akata\inst{2,3}
}
\institute{EECS, University of California, Berkeley CA 94720, USA
\email{\{jinkyu.kim,anna.rohrbach,trevordarrell,canny\}@berkeley.edu}\\ \and 
MPI for Informatics, Saarland Informatics Campus, 66123 Saarbr{\"u}cken, Germany \and 
AMLab, University of Amsterdam, 1098 XH Amsterdam, Netherlands\\
\email{z.akata@uva.nl}}

\maketitle
This supplementary material provides more details on the presented BDD-X dataset and approach as well as more qualitative results. The document as structured as follows. %\anja{do we plan to include a video?}

Section S.1 provides details on our Amazon Mechanical Turk annotation interface and data collection procedure.

Section S.2 includes additional details on the collected descriptions and explanations in BDD-X.

Section S.3 provides implementation details.

Finally, Section S.4 shows more qualitative examples of the predicted visual attention maps as well as textual explanations.

\newpage
\section*{\label{sec:amt}S.1 Our Amazon Mechanical Turk annotation interface}
\begin{figure*}[!h]
    \begin{center}
        \includegraphics[width=\linewidth]{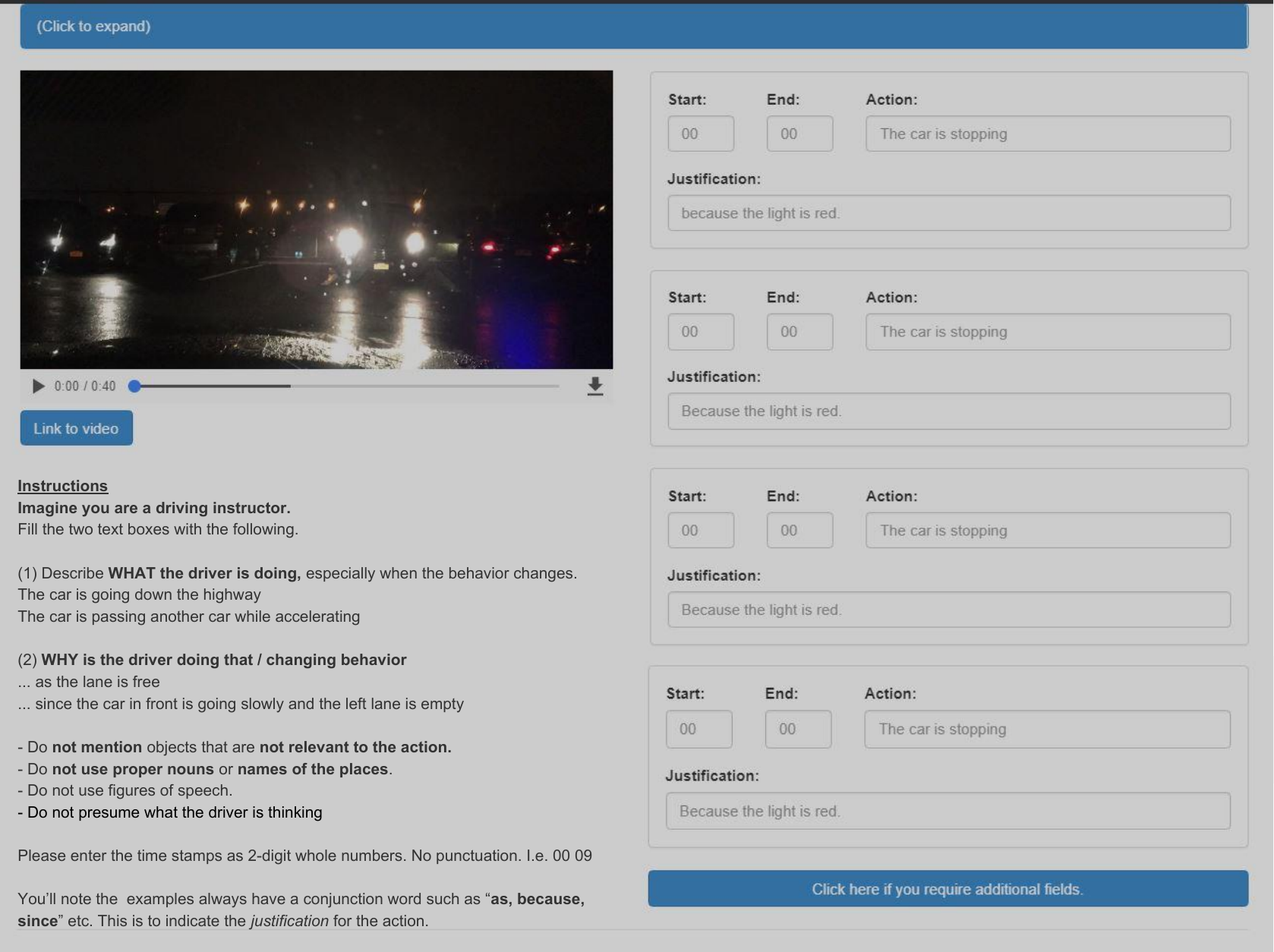}
    \end{center}
    \caption{Our Amazon Mechanical Turk annotation interface. Shown is a ``dummy'' input (The car is stopping + Because the light is red), not an actual annotation.}
    \label{fig:interface}
\end{figure*}

Our annotation prompt is demonstrated on Figure~\ref{fig:interface}. To ensure that the annotators provide us the driving rationales as well as descriptions, we require that they separately enter the action description and the action justification \eg:``The car is moving into the left lane'' and ``because the school bus in front of it is stopping''. In our preliminary annotation studies, we found that giving separate annotation boxes are helpful for the annotator to understand the task and perform better.

\newpage

\section*{\label{sec:dataset}S.2 Berkeley DeepDrive eXplanation (BDD-X) Dataset Details}

\begin{table*}[!b]
\begin{center}
\caption{\label{tbl:stats}BDD-X dataset details. We provide counts for top-30 words, where words are counted along with all their forms, such as \emph{slow, slows, slowing, slowed, slowly}, etc.}

\begin{tabular}{@{}c@{\ \ \ \ \ \ \ \ }c@{}}
\begin{tabular}{@{}lr@{}}
\toprule
\multicolumn{2}{c}{BDD-X action descriptions} \\ 
\midrule
Word & Count \\
\midrule
stop & 6879 \\
slow & 6122 \\
forward & 4322 \\
drive & 3994 \\
move & 3273 \\
accelerate & 2882 \\
right & 2616 \\
left & 2574 \\
turn & 1912 \\
road & 1907 \\
lane & 1832 \\
street & 1704 \\
speed & 1072 \\
come & 1033 \\
merge & 923 \\
pull & 723 \\
intersection & 717 \\
head & 629 \\
continue & 625 \\
slight & 554 \\
make & 539 \\
brake & 517 \\
travel & 513 \\
highway & 450 \\
maintain & 390 \\
steer & 371 \\
proceed & 359 \\
steady & 329 \\
complete & 323 \\
veer & 214 \\
\bottomrule
\end{tabular}
&
\begin{tabular}{@{}lr@{}}
\toprule
\multicolumn{2}{c}{BDD-X action explanations} \\ 
\midrule
Word & Count \\
\midrule
traffic & 7486 \\
light & 6116 \\
red & 3979 \\
move & 3915 \\
clear & 3660 \\
ahead & 3629 \\
road & 3528 \\
stop & 3430 \\
lane & 3407 \\
turn & 3333 \\
front & 2715 \\
green & 1928 \\
park & 1523 \\
intersection & 1513 \\
right & 1464 \\
slow & 1395 \\
cars & 1390 \\
left & 1272 \\
street & 1225 \\
forward & 984 \\
sign & 984 \\
make & 718 \\
approach & 702 \\
speed & 681 \\
down & 658 \\
pedestrian & 649 \\
go & 614 \\
cross & 588 \\
get & 536 \\
pass & 490 \\
\bottomrule
\end{tabular}
\\
\end{tabular}
\end{center}
\end{table*}

Table \ref{tbl:stats} includes word counts for the top-30 most frequent words (excluding stop-words) used in the action descriptions and action explanations, respectively. Note, that word counts are obtained but taking all word forms into account (\emph{slow, slows, slowing, slowed, slowly}, etc). Most common actions are related to changes in speed, driving forward and turning. However many also include merging, pulling, changing lanes, veering, and less frequent actions like reversing, parking or using wipers. Action explanations cover a diverse list of concepts relevant to driving scenario, such as state of traffic/lanes, traffic lights/signs, pedestrians crossing the street, passing other cars, etc. Although less frequent, many explanations contain references to weather conditions (rain, snow), different types of vehicles (buses, vans, trucks), road bumps and safety of the performed action.

We provide some additional examples in Table~\ref{tab:more_examples}, showing, in particular, that some explanations require complex reasoning (1,2,3) and illustrate attention to detail (4,5,6).

% allow resizing
{
\setlength{\tabcolsep}{0.3em}
\renewcommand{\arraystretch}{1} 
\begin{table}[h]
	\begin{center}
    	\resizebox{\linewidth}{!}{%
    	\begin{tabular}{@{}ll@{}} \toprule
        	\footnotesize{} Description  & \footnotesize{} Explanation \\\midrule
            1: The car is making its way carefully through the intersection & because another car has cut it off and pedestrians are approaching the crosswalk.\\
        	2: The car stops behind the truck & because it's waiting for a car in the opposite lane to pass by.\\
            3: The car slows to a stop next to a line of parked cars & since there is no where available to park, but another car up ahead is also double parked.\\
            4:  The car slows down & because it's entering a school crossing zone.\\
            5: The car pulls over to the left side of the street & to avoid a large pothole on the right.\\
            6: The car is moving at a constant slow pace & because it is a single lane road with snow and ice on the road.\\ 
\bottomrule
        \end{tabular}}
        \caption{\label{tab:more_examples} Example annotations where explanations require complex reasoning (1,2,3) and demonstrate attention to detail (4,5,6).}
    \end{center}
\end{table}
}

\newpage
\section*{\label{sec:lstm}S.3 Implementation Details}
\myparagraph{Preprocessing}
Our model is trained to predict two vehicle control commands, \ie an acceleration and a change of course. At each time $t$, a change of course, $c_t$, is computed by taking a difference between a current vehicle’s course $r_t$ and a smoothed value $\bar{r}_t$ by using simple exponential smoothing method~\cite{hyndman2008forecasting}. 
\begin{equation}
c_t = r_t - \bar{r}_t = r_t - \big(\alpha_s r_t + (1-\alpha_s) \bar{r}_{t+1}\big)
\end{equation}
where $\alpha_S$ is a smoothing factor $0\leq\alpha_s\leq1$. Note that they are same as the original timeseries when $s$~=1, while values of $s$ closer to zero have a greater smoothing effect and are less response to recent changes. In this paper, we use $\alpha_s$ as 0.01.  

\myparagraph{Long Short-Term Memory (LSTM) network:}
We use a long short-term memory (LSTM) network~\cite{hochreiter1997lstm} in our vehicle controller and explanation generator. The LSTM is defined as follows:
\begin{equation}
\begin{pmatrix}
{\bf{i}}_t \\
{\bf{f}}_t \\
{\bf{o}}_t \\
{\bf{g}}_t
\end{pmatrix}
=
\begin{pmatrix}
\textit{sigm} \\
\textit{sigm} \\
\textit{sigm} \\
\textit{tanh} 
\end{pmatrix}
\bf{A}
\begin{pmatrix}
{\bf{h}}_{t-1} \\
{\bf{y}}_t
\end{pmatrix}
\end{equation}
where ${\bf{i}}_t$, ${\bf{f}}_t$, ${\bf{o}}_t$, and ${\bf{c}}_t$ $\in{\cal{R}}^\textnormal{M}$ are the $\textnormal{M}$-dimensional input, forget, output, memory state of the LSTM at time $t$, respectively. Internal states of the LSTM are computed conditioned on the hidden state ${\bf{h}}_t\in{\cal{R}}^\textnormal{M}$ and an $\alpha$-weighted context vector ${\bf{y}}_t\in{\cal{R}}^{d}$. We use an affine transformation $\bf{A}: {\cal{R}}^{d+\textnormal{M}} \to \cal{R}^{\textnormal{4M}}$. The logistic sigmoid activation function and the hyperbolic tangent activation function are represented as $\textit{sigm}$ and $\textit{tanh}$, respectively. The hidden state ${\bf{h}}_t$ and the cell state ${\bf{c}}_t$ of the LSTM are defined as:
\begin{equation}
\begin{split}
{\bf{c}}_t &= {\bf{f}}_t \odot {\bf{c}}_{t-1} + {\bf{i}}_t \odot {\bf{g}}_t \\
{\bf{h}}_t &= {\bf{o}}_t \odot \textit{tanh}({\bf{c}}_t)
\end{split}
\end{equation}
where $\odot$ is element-wise multiplication. 

To initialize memory state $c_t$ and hidden state ${\bf{h}}_t$ of the LSTM, we use average of the convolutional feature slices ${\bf{x}}_{0,i}\in{\cal{R}}^{d}$ for $i\in\{0,1,\dots,l\}$ and feed through two additional hidden layers: $f_{\textnormal{init},c}$ and $f_{\textnormal{init},h}$.
\begin{equation}
{\bf{c}}_{0} = f_{\textnormal{init},c}\left(\frac{1}{l}\sum^{l}_{i=1}{\bf{x}}_{0,i}\right),~~
{\bf{h}}_{0} = f_{\textnormal{init},h}\left(\frac{1}{l}\sum^{l}_{i=1}{\bf{x}}_{0,i}\right)
\end{equation}

\newpage
\section*{\label{sec:examples}S.4 Additional Examples}
In this section, we provide more examples of explanations of the driving decisions by exploiting both attention visualization and textual justification.

\begin{figure*}[!h]
    \begin{center}
        \includegraphics[width=0.9\linewidth]{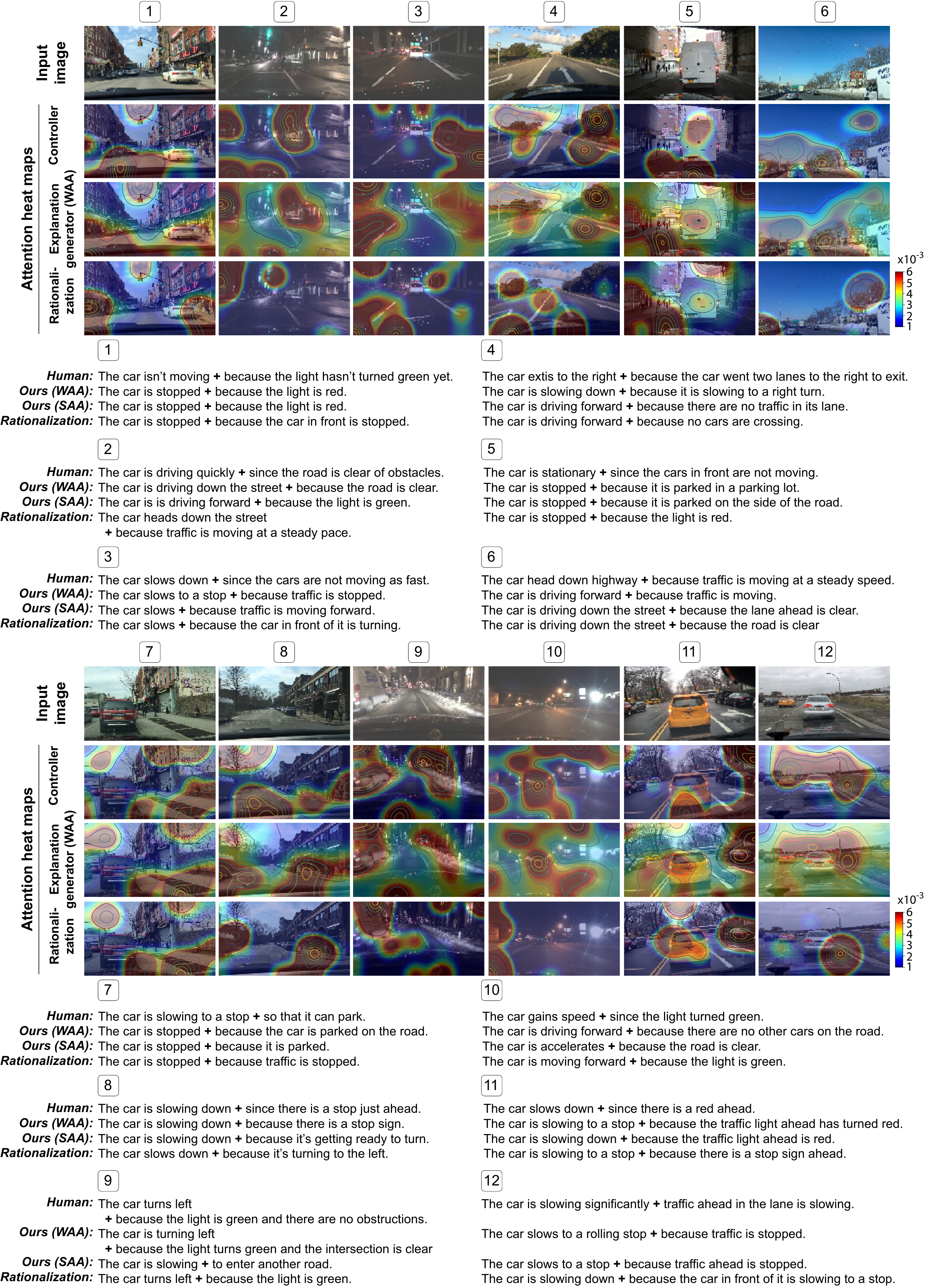}
    \end{center}
    \caption{Examples of descriptions and explanations. In this experiment, we use ($\lambda_c$,$\lambda_a$) as (100,10). A synthetic separator token is replaced by '+'.}
    \label{fig:examples}
\end{figure*}

\begin{figure*}[!h]
    \begin{center}
        \includegraphics[width=0.9\linewidth]{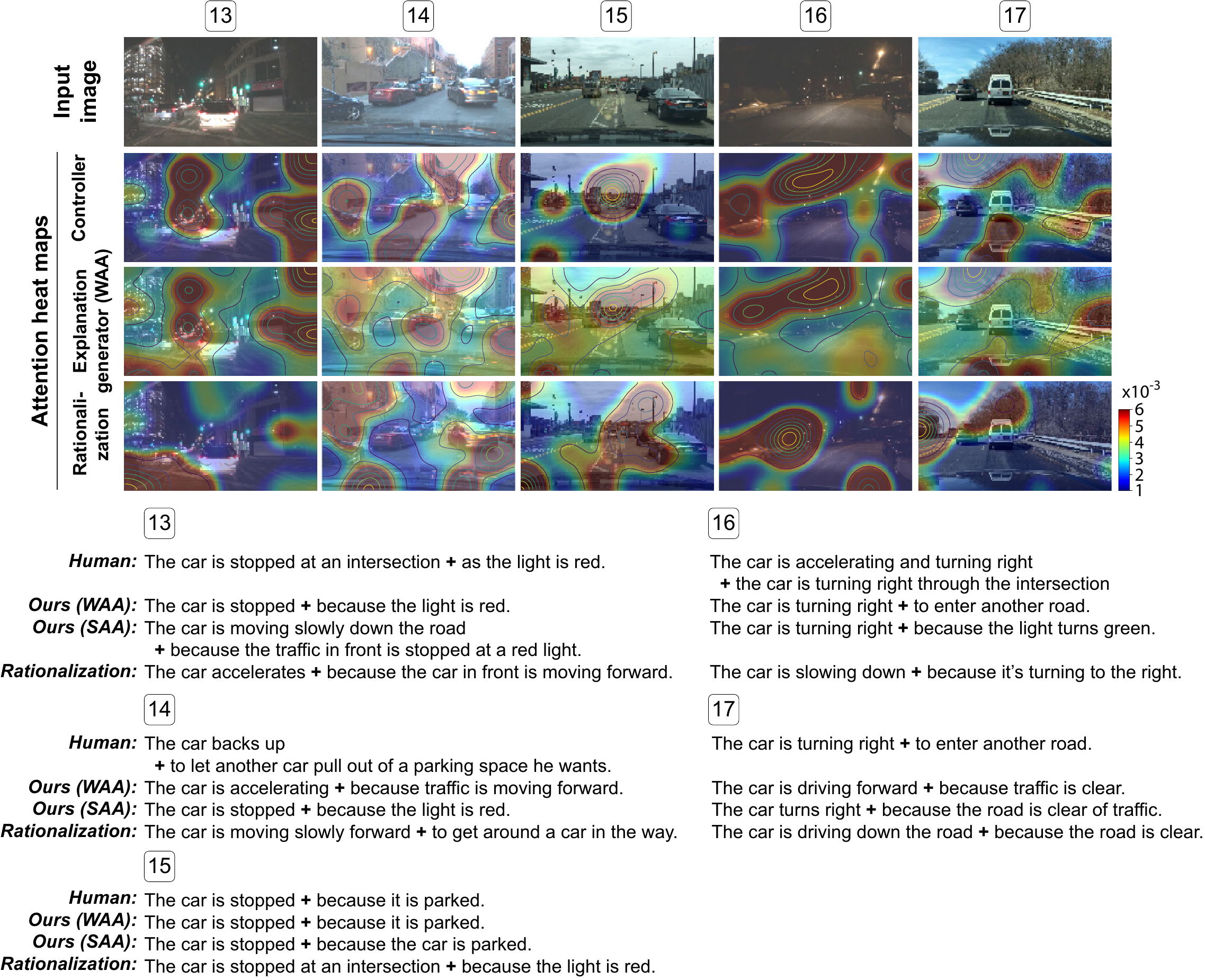}
    \end{center}
    \caption{Additional examples of descriptions and explanations. In this experiment, we use ($\lambda_c$,$\lambda_a$) as (100,10). A synthetic separator token is replaced by '+'.}
    \label{fig:examples}
\end{figure*}

Our model is also able to generate novel explanations not present in the training set. We provide some examples as follows: (1) ``because there are no obstructions in the lane ahead'', (2) ``because the car ahead is slowing to make a stop'', (3) ``because the car is turning onto a different street'', (4) ``as it is now making a turn to enter another street'', (5) ``the car is stopped at an intersection at a red light'', and (6) ``because there are no other cars in the intersection''.

\end{document}